\pdfoutput=1

\documentclass[11pt]{article}

\usepackage{ACL2023}

\usepackage{times}
\usepackage{latexsym}

\usepackage[T1]{fontenc}

\usepackage[utf8]{inputenc}

\usepackage{microtype}

\usepackage{inconsolata}

\usepackage{amsfonts}
\usepackage{amsmath}
\usepackage{pgf-pie}
\usepackage{multirow}
\usepackage{enumerate}
\usepackage{graphicx} 
\usepackage{float}
\usepackage{subfigure}
\usepackage{booktabs}
\usepackage{enumitem}
\usepackage{soul}

\title{DLUE: Benchmarking Document Language Understanding}

\author{Ruoxi Xu${}^{1,3}$,
        Hongyu Lin${}^{1}$\thanks{~ Corresponding Authors},
        Xinyan Guan${}^{1}$,
        Xianpei Han${}^{1, 2}$, \\
        \textbf{Yingfei Sun${}^{3}$,
        Le Sun${}^{1,2}$ } \\
        ${}^{1}$Chinese Information Processing Laboratory ~ 
        ${}^{2}$State Key Laboratory of Computer Science \\
Institute of Software, Chinese Academy of Sciences, Beijing, China\\
        ${}^{3}$School of Electronic, Electrical and Communication Engineering, \\ University of Chinese Academy of Sciences, Beijing, China \\
{\tt \{ruoxi2021,hongyu,xianpei,sunle\}@iscas.ac.cn} \\
\tt yfsun@ucas.ac.cn}

\begin{document}
\maketitle
\begin{abstract}
Understanding documents is central to many real-world tasks but remains a challenging topic. 
Unfortunately, there is no well-established consensus on how to comprehensively evaluate document understanding abilities, which significantly hinders the fair comparison and measuring the progress of the field. 
To benchmark document understanding researches, this paper summarizes four representative abilities, i.e., document classification, document structural analysis, document information extraction, and document transcription.
Under the new evaluation framework, we propose  \textbf{Document Language Understanding Evaluation} -- \textbf{DLUE}, a new task suite which covers a wide-range of tasks in various forms, domains and document genres.
We also systematically evaluate six well-established transformer models on DLUE, and find that due to the lengthy content, complicated underlying structure and dispersed knowledge, document understanding is still far from being solved, and currently there is no neural architecture that dominates all tasks, raising requirements for a universal document understanding architecture.
\end{abstract}

\section{Introduction}

Documents are basic units of the organization of natural language~\citep{buckland1997document}.
Understanding the structures and the semantics of documents is the foundation for understanding news articles~\citep{kiesel2019semeval}, scientific papers~\citep{dasigi-etal-2021-dataset}, government reports~\citep{huang-etal-2021-efficient} , stories~\citep{kovcisky2018narrativeqa}, etc.
Evaluating how a machine intelligence system can read, analyze and generate documents is an important part of evaluating its natural language abilities, which has long been a critical direction in NLP field.

\begin{figure}
    \centering
    \setlength{\belowcaptionskip}{-0.4cm}
    \includegraphics[width=0.49\textwidth]{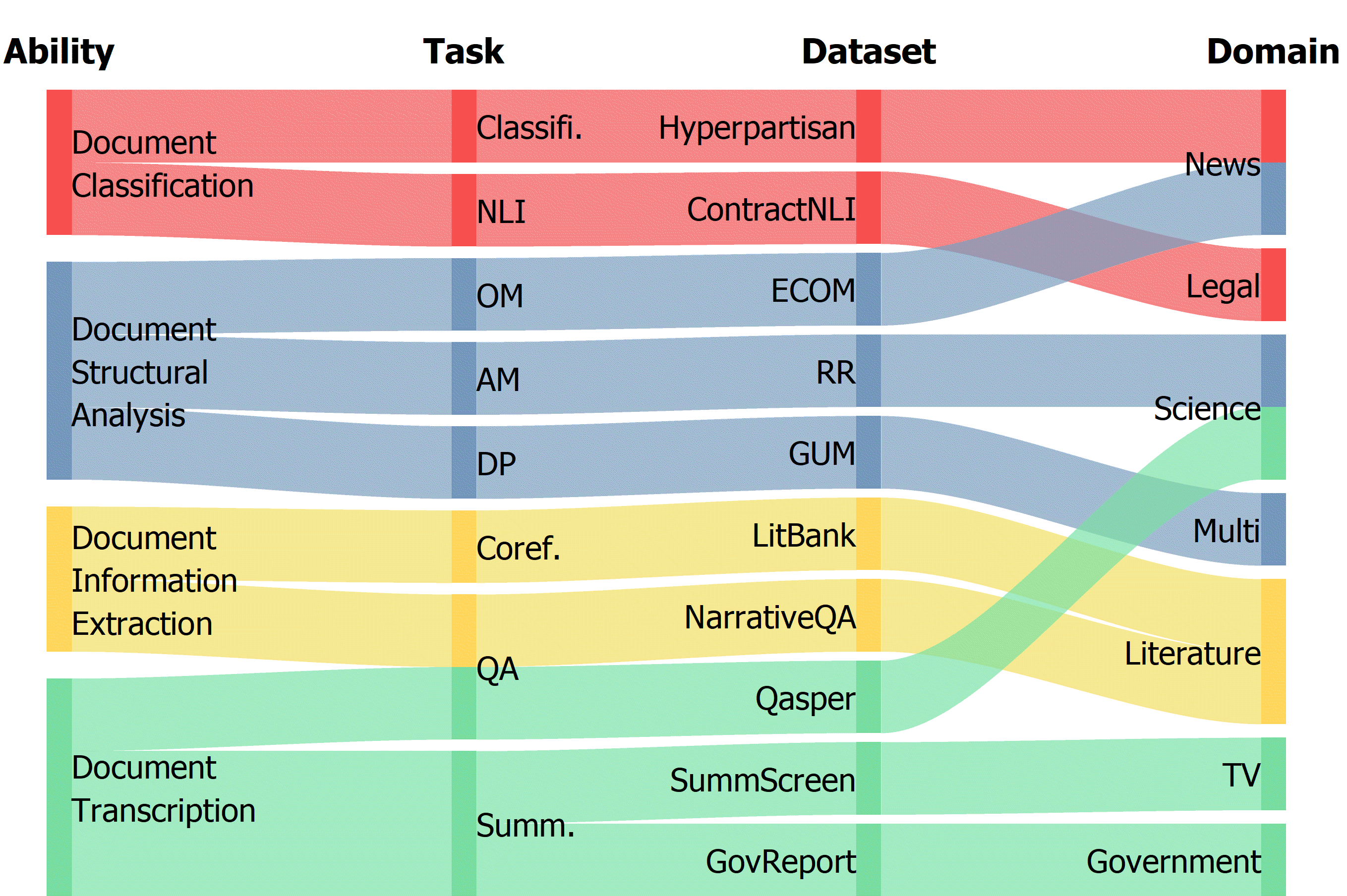} 
    \caption{Overview of DLUE, which covers a wide-range of \textbf{tasks} and \textbf{datasets} in diverse \textbf{domains} to evaluate four representative document understanding \textbf{abilities}.}
    \label{dlue}
\end{figure}

While standard benchmarks like GLUE~\citep{shanahan2016challenge} and SuperGLUE~\citep{wang2019superglue} have become a critical part of NLP community, they primarily focused on short utterances like sentences or paragraphs.
However, documents are much more than bag-of-sentences, which usually focus on the central theme~\citep{benamara2017evaluative}, with underlying structure bound by complex linguistic elements ~\citep{parsing2009speech} and dependent knowledge dispersed across the whole text~\citep{huang2021document}.
Therefore, these benchmarks cannot be used to evaluate document understanding due to its unique challenges:
\emph{First, documents usually have lengthy content}, i.e., they usually are much longer than sentences/paragraphs thus it is difficult to process them due to the computational memory/runtime limitation of current NN models.
\emph{Second, documents have underlying structures}, which play a critical role during understanding. For example, mining arguments from a document needs to model both the local coherence between sentences and the global interactions between claims and arguments, which cannot be accomplished by only exploiting sentence- or paragraph-level information.
\emph{Third, the knowledge in a document is usually dispersed beyond sentences/paragraphs}, which makes it necessary to model and explore document-level context. For example, long-distance within-document coreference resolution needs to integrate global information to find all expressions that refer to the same entity distributed in the full text.

Recently, an increasing number of researches have paid attention to the evaluation of document understanding.
~\citet{woah-2021-online} proposes the Long Range Arena (LRA), which contains two synthetic text-related tasks and evaluates model quality of understanding multi-modal long contexts.
~\citet{hudson2022muld} proposes MuLD, which concentrates on merged sequences over 10,000 tokens. SCROLLS~\citep{shaham2022scrolls} is a recently proposed benchmark that contains NLI, QA and summarization tasks and focuses on long language sequences.
However, these benchmarks mainly focus on the lengthy content challenge, while ignoring other important challenges. As a result, almost all tasks in these benchmarks can be resolved via an retrieval-answering paradigm, i.e., retrieving a very limited number of sentences that contains critical information and then resolving the task. Furthermore, these benchmarks only cover limited tasks, which makes them unable to thoroughly evaluate the document understanding abilities of models.

To systematically evaluate document language understanding abilities, this paper proposes \textbf{Document Language Understanding Evaluation -- DLUE}, a new task suite which covers a wide-range of tasks in various forms, different domains and document genres.
Figure~\ref{dlue} shows the overview of DLUE.
Specifically, we summarize 4 kinds of document understanding abilities, including 1) Document Classification, which evaluates whether a model can understand the overall semantics of a document, e.g., its topic~\citep{zhang2015character} and standpoint~\citep{kiesel2019semeval}; 2) Document Structural Analysis, which evaluates whether a model can analyze and leverage the underlying structure of a document, e.g., its discourse structure~\citep{10.1007/s10579-016-9343-x} and argument structure~\citep{cheng2020ape}; 3) Document Information Extraction, which evaluates whether a model can recognize and aggregate associated information spanning cross whole document, e.g., long-distance within-document coreference~\citep{bamman-etal-2020-annotated}; 4) Document Transcription, which evaluates whether a model can capture and transcript important information of a document, e.g., summarization~\citep{huang-etal-2021-efficient,chen-etal-2022-summscreen} and abstractive QA~\citep{dasigi-etal-2021-dataset}.
Then we collect 10 datasets and align them with the above abilities. Datasets of the same group are converted to a unified format.
In this way, DLUE provides a comprehensive benchmark for document language understanding, which enables the research community to fairly compare and measure the progress of this field.

To better understand the challenges of document language understanding and analyze the performance of current approaches, we conduct experiments on DLUE using several state-of-the-art document understanding models, including 1) Memory-based approaches, which includes XLNet~\citep{yang2019xlnet}; 2) Pattern-based approaches, which includes Longformer~\citep{beltagy2020longformer}, BigBird~\citep{zaheer2020big}, Sparse transformer~\citep{child2019generating}; 3) Low-rank/Kernel-based approaches, which includes Linformer~\citep{wang2020linformer} and Performer~\citep{choromanski2020rethinking}.
Experiments show that document understanding is still far from being solved due to lengthy content, complicated underlying structure and dispersed knowledge, and currently there is no neural architecture that dominates all tasks, raising requirements for a universal document understanding architecture.

Generally, the contributions of this paper are:
\begin{itemize}[leftmargin=18pt,topsep=2pt,itemsep=1pt,parsep=1pt]
    \item We summarize 4 representative abilities for lengthy, structural and global document understanding, including document classification, document structural analysis, document information extraction and document transcription.
    \item We propose a comprehensive benchmark for document language understanding -- DLUE\footnote{\href{www.dluebenchmark.com}{dluebenchmark.com}}, which is built on established annotated datasets and selected to cover a diverse range of text genres, dataset sizes, and degrees of difficulty.
    \item We evaluate current state-of-the-art document understanding models on DLUE, which provides a novel reference to assess current models' abilities and properties when they handle different kinds of document understanding tasks.
\end{itemize}

\section{Background}

\paragraph{NLP Benchmarks}
The development of natural language understanding (NLU) evaluation benchmarks has helped drive the progress of pretraining and transfer learning in NLP.
Benchmarks proposed in the early stage mostly aim at general tasks, such as SentEval~\citep{conneau2018senteval} for universal sentence representations, DecaNLP~\citep{mccann2018natural} for ten diversified NLP tasks cast as a general question-answering format, GLUE~\citep{wang2018glue} for NLU in the English language and SuperGLUE~\citep{wang2019superglue} as a harder counterpart of GLUE.
Besides, benchmarks for more specific tasks have also been proposed, such as DialoGLUE~\citep{mehri2020dialoglue} for task-oriented dialogue, DiscoEval~\citep{chen2019evaluation} for discourse-aware sentence representations, GLUECoS~\citep{khanuja2020gluecos} for code-switched NLP, KILT~\citep{petroni2020kilt} for knowledge-intensive language tasks, and etc.

The above benchmarks mostly focus on sentences or short texts. However, documents are also very common for complex tasks or real-world textual data. 
Single-task benchmark for document understanding mostly uses summarization tasks~\citep{cohan2018discourse} or QA tasks~\citep{dasigi-etal-2021-dataset}. Due to the single nature of the task and data distribution, it is difficult for these benchmarks to comprehensively evaluate models' ability to model documents.
There are also some multi-task benchmarks for document understanding, such as the Long Range Arena (LRA)~\citep{tay2020long}, SCROLLS~\citep{shaham2022scrolls}, MuLD~\citep{hudson2022muld} and LOT~\citep{guan2022lot}. Long inputs of LRA and MuLD are either automatically generated or artificially lengthened. Tasks in SCROLLS all focus on a few sentences or paragraphs, which can be solved by retrieval-based or chunk-based approaches. LOT only focuses on Chinese long text understanding and generation. In this paper, compared with existing benchmarks that focus on long sequences instead of documents, we focus on challenges posed by document understanding, including lengthy content, complicated underlying structure and dispersed knowledge.

\paragraph{Document Understanding Models}
There have been numerous attempts to improve both the memory footprint and computational cost of transformers, thus allowing the use of longer inputs. 
A natural way is to connect blocks via recurrence, such as XLNet~\citep{yang2019xlnet}.
Another way of tackling the high complexity of full attention is to sparsify the attention matrix. Longformer~\citep{beltagy2020longformer}, BigBird~\citep{zaheer2020big} and Sparse transformer~\citep{child2019generating} simply sparsify the attention matrix by limiting the field of view to fixed, predefined patterns such as local windows and block patterns of fixed strides. Reformer~\citep{kitaev2020reformer} uses learnable ones, an extension to fixed, pre-determined pattern.
Besides, low-rank approximations or kernelization of the self-attention matrix can be used as well to reduce the complexity. Linformer~\citep{wang2020linformer} and Performer~\citep{choromanski2020rethinking} are representative low-rank/kernel-based transformers.
In this paper, we conduct experiments on the above three document understanding architectures to explore challenges posed by document understanding.

\section{DLUE: Document Language Understanding Evaluation  Benchmark}
\label{section_dlue}

\begin{table}[!t]
    \centering
    \setlength{\belowcaptionskip}{-10pt}
    \resizebox{0.5\textwidth}{!}{
        \begin{tabular}{ccccccc}
            \toprule
            \multirow{2}*{Corpus} & \multirow{2}*{Task} & \multirow{2}*{Domain} & \multirow{2}*{Metric} & \multicolumn{2}{c}{Avg \#Words} & \multirow{2}*{\#Examples} \\
            & & & & Input & Output & \\
            \midrule
            \multicolumn{7}{c}{Classification} \\
            Hyperpartisan & Classifi. & News & acc. & 588 & 1 & 1273 \\
            ContractNLI & NLI & Legal & acc. & 1708 & 1 & 10319 \\
            \midrule
            \multicolumn{7}{c}{Structure Analysis} \\
            ECOM & OP & News & $F_1$ & 488 & 20 & 2000 \\
            RR & AM & Science & $F_1$ & 793 & 47 & 4764 \\
            GUM & DP & Multi & $F_1$ & 939 & 119 & 175 \\
            \midrule
            \multicolumn{7}{c}{Extraction} \\
            LitBank & Coref. & Literature & $F_1$ & 2115 & 7.4 & 7214 \\
            NarrativeQA & QA & Literature & $F_1$ & 51790 & 4.6 & 71187 \\
            \midrule
            \multicolumn{7}{c}{Transcription} \\
            GovReport & Summ. & Government & ROUGE & 7897 & 492.7 & 19402 \\
            SummScreen & Summ. & TV & ROUGE & 5639 & 100.0 & 4348 \\
            Qasper & QA & Science & $F_1$ & 3671 & 11.5 & 5692 \\
            \bottomrule
        \end{tabular}
    }
    \caption{Task descriptions and statistics of DLUE.}
    \label{dataset}
\end{table}

The section describes the DLUE benchmark, which is used to evaluate the 4 representative abilities of document understanding. Specifically, DLUE is centered on 10 English document understanding datasets, which cover a wide-range of tasks in various forms, different domains and document genres. Table~\ref{dataset} provides an overview of the datasets included in the benchmark. In the following, we describe the details of DLUE.

\subsection{Overview}
As described above, a document understanding system should resolve the lengthy content, complicated underlying structure, and dispersed knowledge challenges.
To effectively evaluate the above abilities and challenges, DLUE selects datasets according to the following several desiderata:
First, the documents in the benchmark should have lengthy content. We select datasets with an average token number of more than 512, considering the fact that most existing state-of-the-art NLP systems (e.g., pretrained models) are limited to 512 to 1024 tokens~\citep{devlin2018bert}.
Second, the tasks must be solved using the dispersed knowledge in a document. Therefore, we don't select a document-level dataset if most of them can be resolved through chunk-based or retrieval-based approaches. 
Third, the documents in the benchmark must be natural, such as literature works, scientific articles, government reports and so on. Synthesized documents don't have structure information and relation links among different sections.
Fourth, the selected tasks should be beyond the scope of current state-of-the-art systems, but solvable by most college-educated English speakers.

Based on the above desiderata and with the permission of licenses, we collect as diverse datasets as possible to increase the coverage on capabilities.
The overview of DLUE is shown in Figure \ref{dlue}, and their statistics are shown in Table 1. In the following, we describe all datasets according to the their target ability.

\subsection{Document Classification}
A document usually narrow focus on a single central theme~\citep{benamara2017evaluative}.
We aim to evaluate document classification ability, specifically the ability to understand the overall semantics of documents in this section.
To do this, we select two datasets that rely on full-text to make judgements and reformulate every dataset as document classification tasks. Specifically, given single sequence $s$ or sequence pairs $(s_1,s_2)$, the goal is to classify the input into a single label $l$.

\paragraph{Hyperpartisan} ~\citep{kiesel2019semeval} is a document classification dataset which aims to automatically detect news that takes an left-wing or right-wing standpoint.
A few words or sentences are not enough to determine the political leanings of news, which are toned by the full text.
This task provides two datasets, one is labeled manually and the other is labeled in a semi-automated manner via distant supervision at the publisher level. We use the first one to pursue higher evaluation accuracy and keep the same train/test split as the original work. 

\paragraph{ContractNLI}~\citep{koreeda2021contractnli} is a natural language inference dataset in the legal domain, with non-disclosure agreements (NDAs) as premises and legal statements as hypothesizes. NDAs are collected from Internet search engines and Electronic Data Gathering, Analysis, and Retrieval system (EDGAR). 
To correctly predict whether the hypothesis is entailed, neutral, or contradictory from the contract, we need to refer to not necessarily continuous sentences across the contract with hundreds of tokens.
The dataset contains 607 contracts and 17 unique hypothesizes, which we combine to produce 10319 instances.

\subsection{Document Structure Analysis}
A document is composed of structured groups of sentences, paragraphs and sections. Analyzing document structure can be very useful in indexing and organizing the information contained in the document.
Tasks in this section aim to evaluate document structure analysis ability, specifically the ability to capture and leverage structure information. 
We select three datasets as follows and reformulate every dataset as sentence-level sequence labeling tasks. Specifically, given a document $d=\{s_1,s_2,...,s_n\}$, the goal is to output a tag sequence $t=\{t_1,t_2,...,t_n\}$ for sentences.

\paragraph{ECOM}~\citep{xu2022eco} is an event-centric opinion mining corpus in which a system takes in an event descriptor and related news articles to extract event-related opinion segments from articles.
An opinion segment is composed of continuous sentences targeting at the same argument. We select the dataset to evaluate the ability to utilize 
local structure information, which is important for identifying opinion boundaries unambiguously.

\paragraph{RR}~\citep{cheng2020ape} is an argument mining corpus for extracting arguments from reviews and rebuttals, which are collected from ICLR 2013 - 2020 (except for 2015 that is unavailable) from openreview.net.
Peer reviews and rebuttals on scientific works are a data source of rich structures and long passages.
We think it's a suitable dataset because experiments in the original paper show that the internal structure information is important for this task.

\paragraph{GUM}~\citep{10.1007/s10579-016-9343-x} is a multi-layer corpus collected and edited via classroom annotation. We focus on its Rhetorical Structure Theory analysis annotation.
We consider the task of predicting annotated discourse relations among sentences, as it's the most direct way to probe structure knowledge. The problem is framed to a sequence labeling task as ~\citet{koto2021top}, where the goal is to iteratively find a segmentation boundary to split a sequence of discourse units into two sub-sequences of discourse units.

\subsection{Document Information Extraction}
Dependent knowledge in a document is usually dispersed across the full text, which plays an important role in the transmission of the intended meaning.
Tasks in this section aim to evaluate document information extraction ability, specifically the ability to identify long-distance related mentions and relations. 
We select two datasets as follows and reformulate every dataset as multi-answer question answering tasks. Specifically, given a document $d$ and a question $q$, the goal is to extract correct answer spans $a=\{a_1, a_2, ...,a_n\}$ from $d$ for $q$.

\paragraph{LitBank}~\citep{bamman-etal-2020-annotated} is a coreference resolution dataset on English literature works.
The documents in LitBank are several times longer than those in other benchmark datasets (e.g. 463.7 tokens for OntoNotes) and thus are abundant with long-distance within-document coreference.
For each coreference link, we transform the sentence of one mention into a question, take all mentions as answers, and then can get 7214 question-answer pairs.

\paragraph{NarrativeQA}~\citep{kovcisky2018narrativeqa} is a reading comprehension dataset on books and movie scripts.
The questions in NarrativeQA are written based on summaries. Therefore, whether to understand or answer questions requires to recognize long-distance associated information according to several parts or a larger span of the context document.

\subsection{Document Transcription}
Tasks in this section aim to evaluate document transcription ability, specifically the ability to capture and transcript key information of documents.
We select three datasets that need to contextualize across different sections and reformulate every dataset as sequence-to-sequence tasks. Specifically, given a sequence $s$, the goal is to output a concise
and fluent new sequence $s_N$.

\paragraph{GOVREPORT}~\citep{huang-etal-2021-efficient} is a summarization dataset of long reports on various national policy issues and paired expert-written summaries published by U.S. Government Accountability Office (GAO) and Congressional Research Service (CRS).
Documents and summaries in GovReport are significantly longer than prior datasets, such as 1.5 times longer than Arxiv~\citep{cohan-etal-2018-discourse}.
Moreover, new salient bigrams are steadily added as more content is consumed, which indicates information is spread throughout documents in the dataset.

\paragraph{SummScreen}~\citep{chen-etal-2022-summscreen} is a summarization dataset comprised of pairs of TV series transcripts and human written recaps.
Different from official documentation like GOVREPORT~\citep{huang-etal-2021-efficient}, in this dataset, the language expression is more informal and the structure is more unclear. We need to combine the whole document to understand plots that are often expressed indirectly in character dialogues and scattered across the entirety of the transcript.

\paragraph{Qasper}~\citep{dasigi-etal-2021-dataset} is a QA dataset in the research domain focusing on entire papers, in which both questions and answers are handed-written by NLP practitioners. Over half of questions require multiple paragraphs as evidence to answer. We prepend the query to the document, using two newlines as a natural separator to construct the input sequence.

\begin{table*}[!t]
    \centering
    \setlength{\belowcaptionskip}{-10pt}
    \resizebox{\textwidth}{!}{
        \begin{tabular}{c|cc|cc|ccc|cc|ccc|c|c}
            \toprule
            \multirow{2}*{Model} & \multirow{2}*{\#param} & \multirow{2}*{pretrain} & \multicolumn{2}{c}{Classification} & \multicolumn{3}{c}{Structure Analysis} & \multicolumn{2}{c}{Extraction} &  \multicolumn{3}{c}{Transcription} & \multirow{2}*{Avg} & Inference Speed\\
            & & & Hyper & CNLI & ECOM & RR & GUM & LitBank & NrQA & SummScr & GovRep & Qasper & & (steps per sec)\\
             \midrule
            & & & \multicolumn{11}{c}{Vanilla Transformer} \\
            \midrule
            BERT & 110M & yes & 80.1 & 72.3 & 37.3 & 57.3 & - & 34.1 & 14.5 & - & - & - & - & -\\
            \midrule
            & & & \multicolumn{11}{c}{Memory-based} \\
            \midrule
            XLNet & 110M & yes &81.4 &80.2 &\textbf{39.1} &\textbf{74.0} &65.4 &78.1 &15.2 &18.9 &22.7 &24.2 & 42.8 & 0.95\\
            \midrule
            & & & \multicolumn{11}{c}{Pattern-based} \\
            \midrule
            Longformer & 148M & yes &83.8 &71.6 &37.9 &72.9 &58.4 &\textbf{79.1} &\textbf{18.3} &\textbf{20.9} &25.7 &\textbf{26.4}& 43.2 & 2.0\\
            BigBird & 127M & yes  &\textbf{85.9} &\textbf{82.8} &37.0 &71.1 &\textbf{67.6} &77.8 &18.2 &20.6 &\textbf{27.3} &26.2 & \textbf{44.5} & 1.6\\
            Sparse Trans. & 46M & no &64.6 &67.7 &21.9 &45.6 &47.5 &56.7 &11.1 &21.4 &17.6 &17.6 & 31.7 & 3.8\\
            \midrule
            & & & \multicolumn{11}{c}{Low-rank/Kernel-based} \\
            \midrule
            Linformer & 33M & no & 67.1 &65.5	&22.6	&44.3	&53.4 & 63.8 & 12.4 &18.9 &25.8 &17.5 & 32.1 & 6.4\\
            Performer & 51M & no & 67.9 & 69.5 &18.6 &48.6 &56.8 &51.6 &10.1 &20.1 &15.6 &21.5 & 33.0 & 6.7\\
            \bottomrule
        \end{tabular}
    }
    \caption{Overall experimental results on DLUE. Best model is in boldface. "-" denotes the model can't handle this task.}
    \label{overall_result}
\end{table*}

\section{Experiments and Analysis}

\subsection{Benchmarking Architectures}

This section describes models and architectures we evaluate on DLUE. Following the general taxonomy of efficient transformer models~\citep{tay2020efficient}, we conduct experiments on six well-established transformer variants to represent a diverse cross-section of document understanding models. Specifically, aside from the standard vanilla transformer, we compare three approaches:

\begin{itemize}
    \item Memory-based models, including XLNet~\citep{yang2019xlnet}.
    \item Pattern-based models, including Longformer~\citep{beltagy2020longformer}, BigBird~\citep{zaheer2020big} and Sparse Transformer~\citep{child2019generating}.
    \item Low-rank/Kernel-based models, including Linformer~\citep{wang2020linformer} and Performer~\citep{choromanski2020rethinking}.
\end{itemize}

For all kinds of task formulations described in Section ~\ref{section_dlue}, we implement unified model architectures. For document classification tasks, we use the special classification token ([CLS]) for prediction. Specifically, we concatenate a [CLS] token in front of each sequence and then input them into encoders. The final hidden vector of [CLS] token is taken as the aggregate representation and passed into a two-layered MLP with ReLU activations for classification. The document structure analysis tasks are reformulated into sentence-level sequence labeling tasks. We use the classical Transformer-CRF architecture as in named entity recognition~\citep{devlin2018bert}. Specifically, we insert external [CLS] tokens at the start of each sentence, and each [CLS] symbol collects features for the sentence preceding it~\citep{liu2019text}. Then the sentence representations are input into Conditional Random Field~\citep{lafferty2001conditional} to get the sentence-level labels. The document information extraction tasks are reformulated into multi-span question answering tasks. Following ~\citet{hu2019multi}, we expand traditional MRC architecture by adding a span number prediction and search component. For transcription tasks, we use the basic encoder-decoder architecture~\citep{vaswani2017attention}. 

\subsection{Implementations}

Our models are implemented by PyTorch framework\footnote{https://pytorch.org/}\footnote{In practice, these models often use a combination of the proposed approximate global attention and simple local attention.}.
For transformers with public pretrained models, we use the base version, including XLNet-base, Longformer-base, BigBird-base. The learning rate is 1e-5 for pretrained models and 1e-3 for classifier heads.
For other models, we follow the same setup as Long range arena~\citep{tay2020long}, a widely recognized benchmark for efficient transformers to minimize the influence of hyper-parameter settings. These transformer models are parameterized by the same number of layers, heads and hidden dimensions, namely 6 layers, 8 heads, 512 hidden dimensions and d = 2048 for positional FFN layers. We use Adam with warmup. All models are trained for 10 epochs. Across datasets and models, we run three repetitions with different random seeds and report averaged scores.

\begin{figure*} [t!]
	\centering
	\setlength{\belowcaptionskip}{-0.4cm}
	\subfigure[Hyperpartisan]{\label{fig:subfig:a}\includegraphics[width=0.24\textwidth]{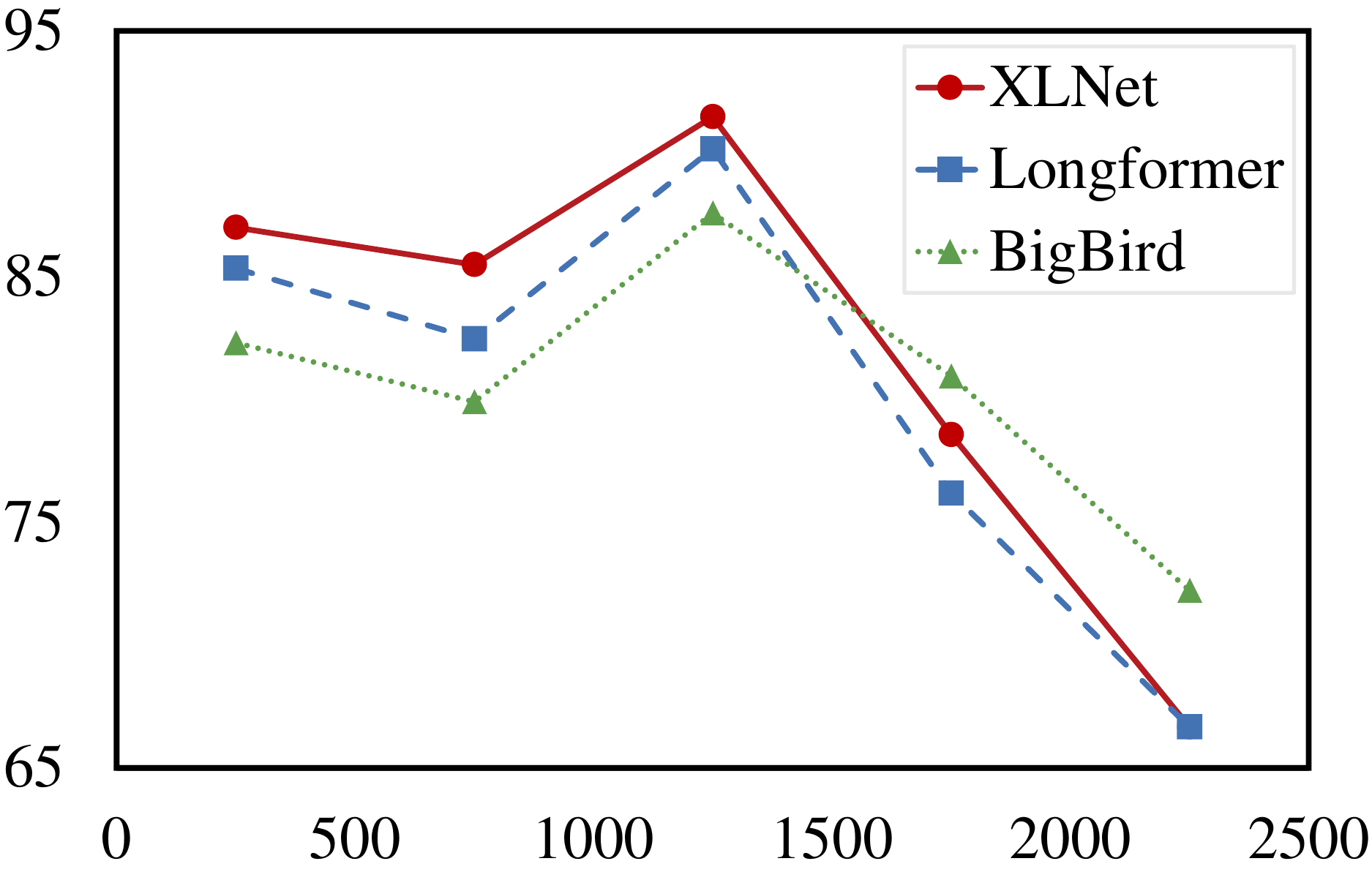}}
	\subfigure[ContractNLI]{\label{fig:subfig:b}\includegraphics[width=0.24\textwidth]{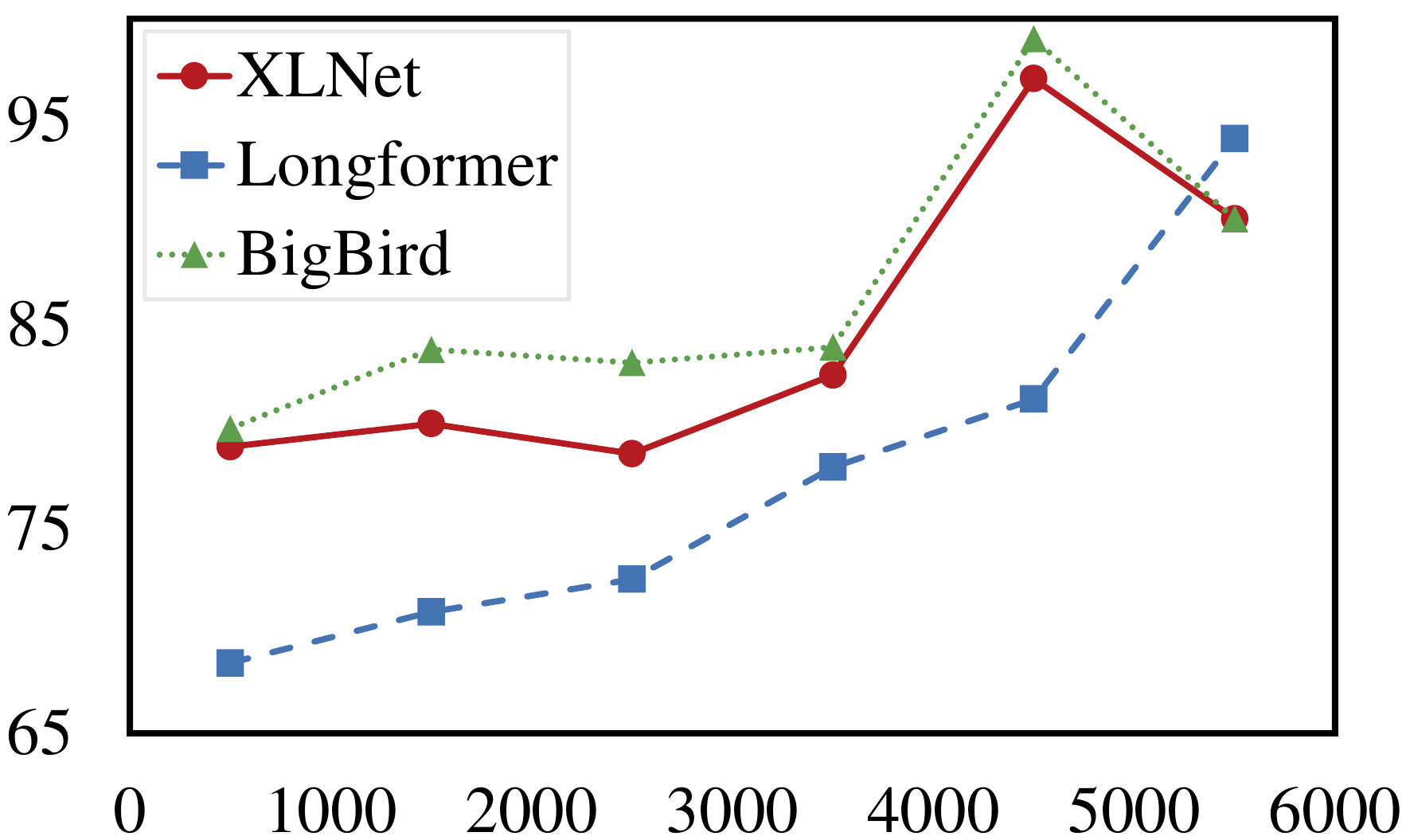}}
	\subfigure[RR]{\label{fig:subfig:c}\includegraphics[width=0.24\textwidth]{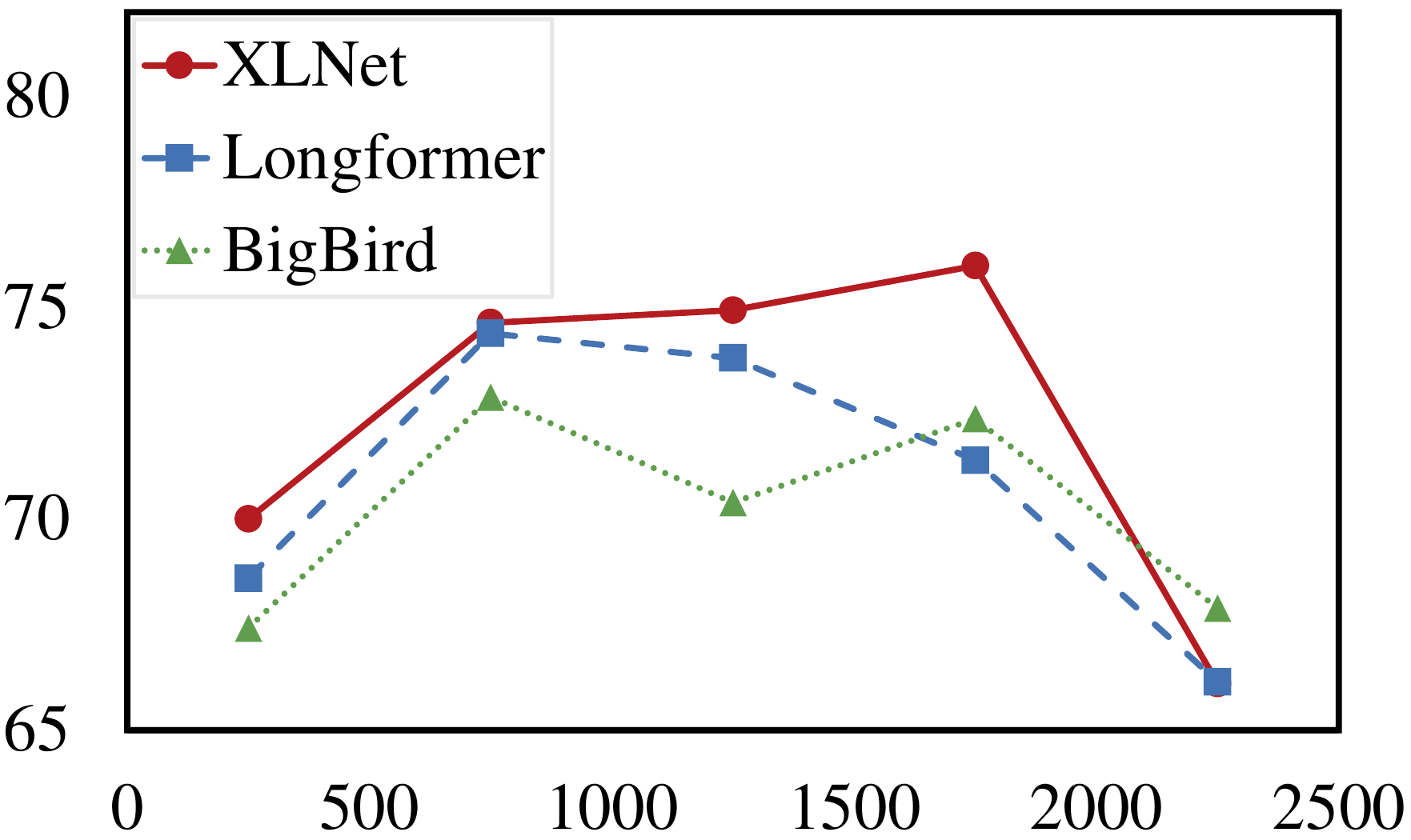}}
	\subfigure[ECOM]{\label{fig:subfig:d}\includegraphics[width=0.24\textwidth]{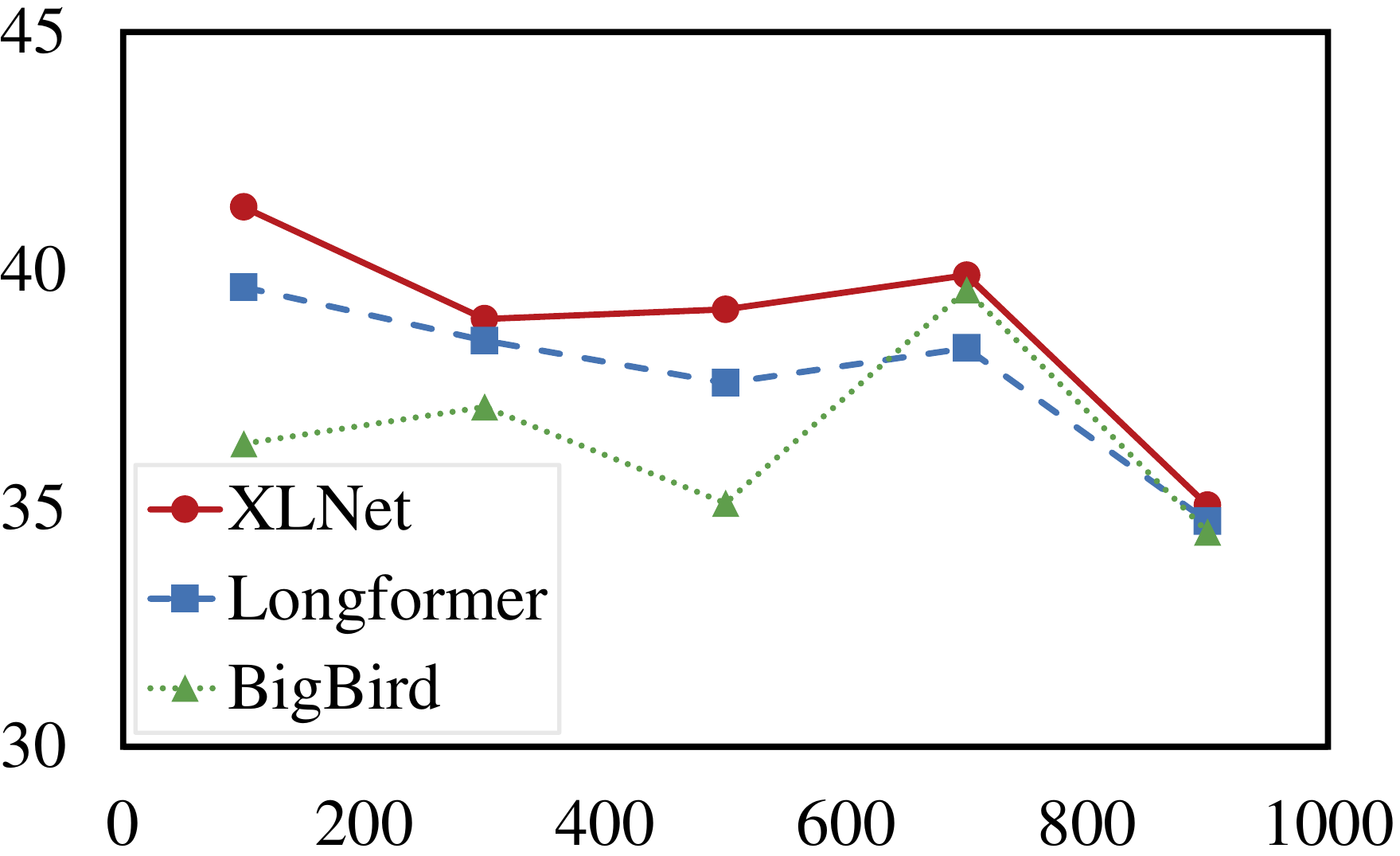}} \\
	\subfigure[LitBank]{\label{fig:subfig:e}\includegraphics[width=0.24\textwidth]{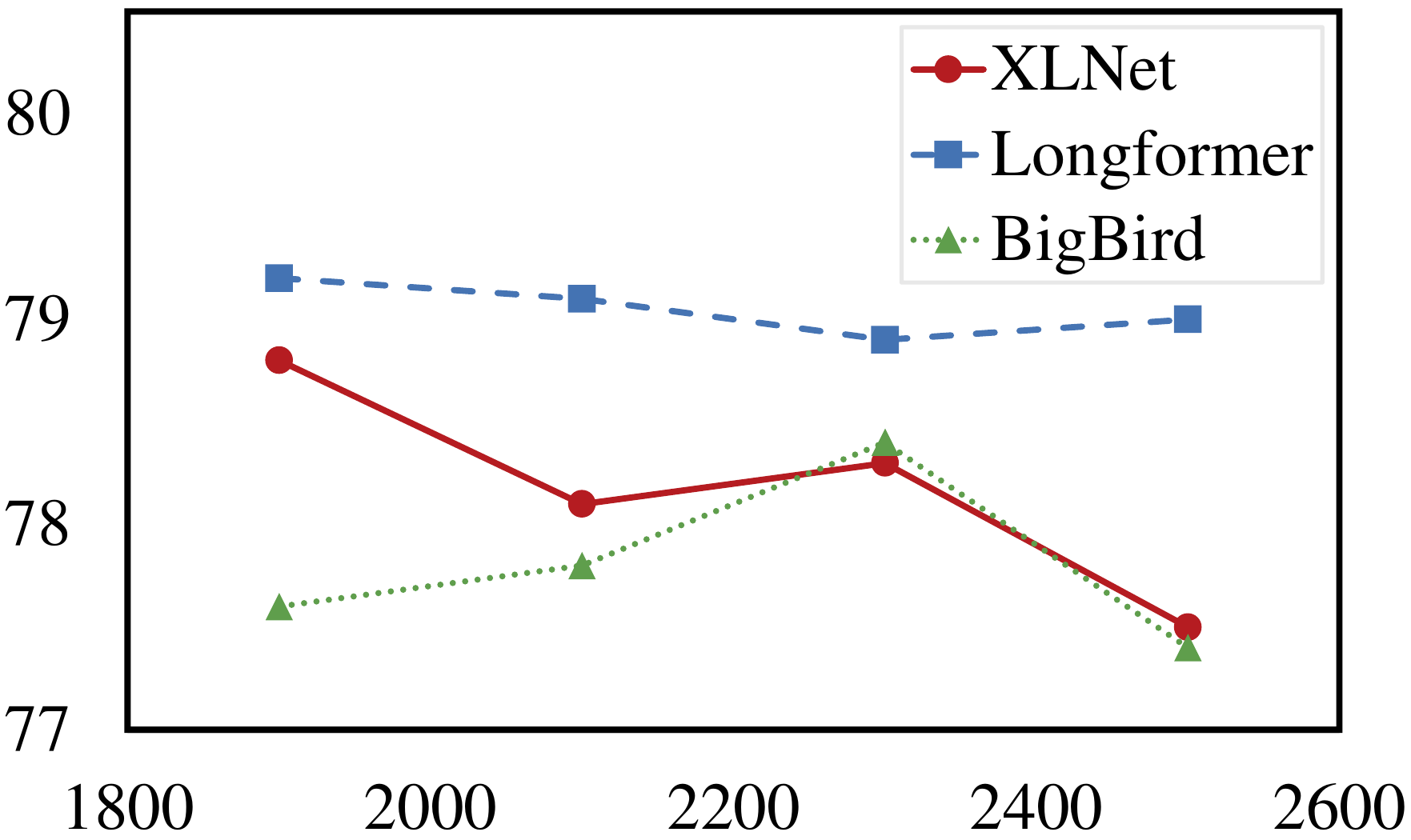}}
	\subfigure[NarrativeQA]{\label{fig:subfig:f}\includegraphics[width=0.24\textwidth]{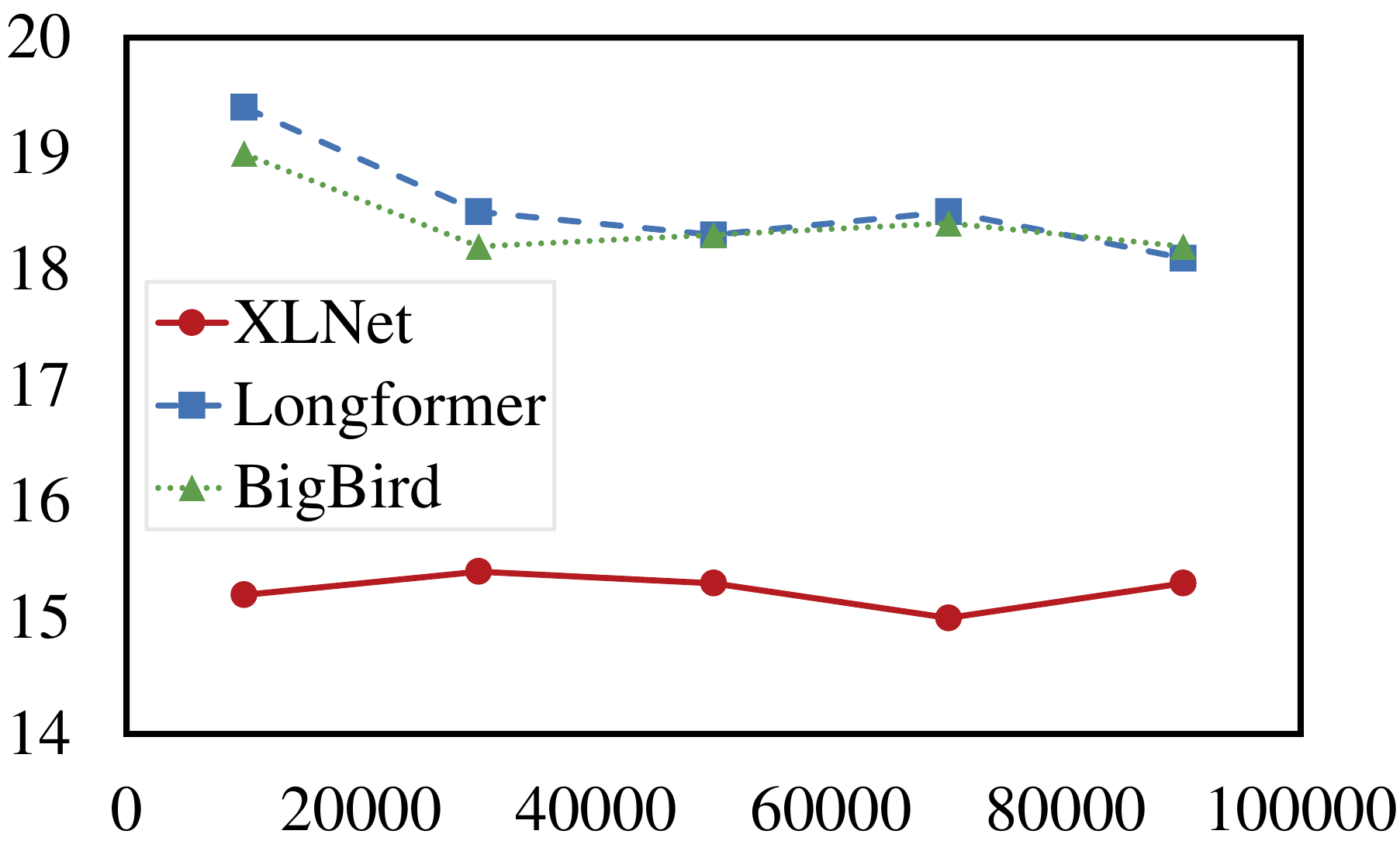}}
	\subfigure[SummScreen]{\label{fig:subfig:g}\includegraphics[width=0.24\textwidth]{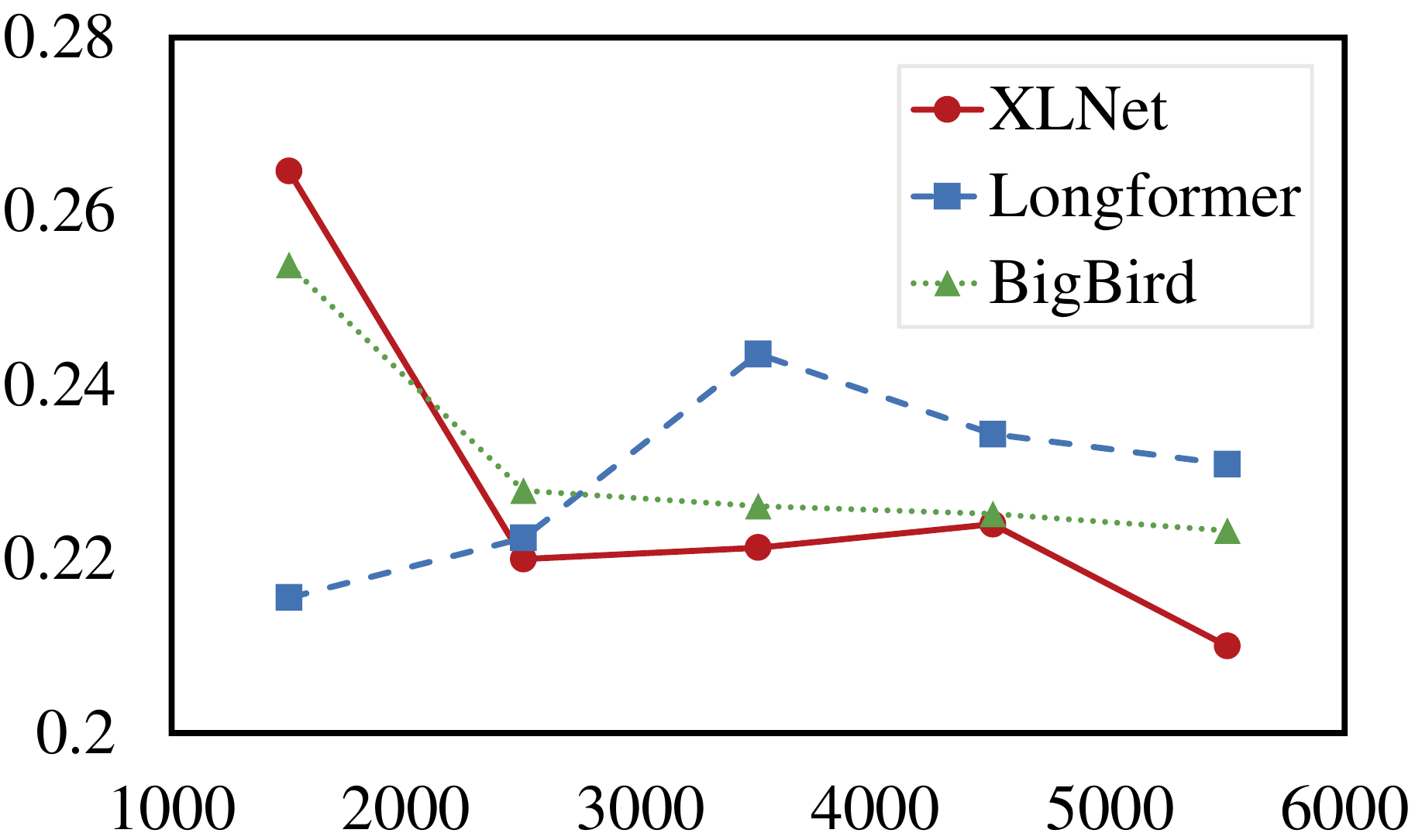}}
	\subfigure[GovReport]{\label{fig:subfig:h}\includegraphics[width=0.24\textwidth]{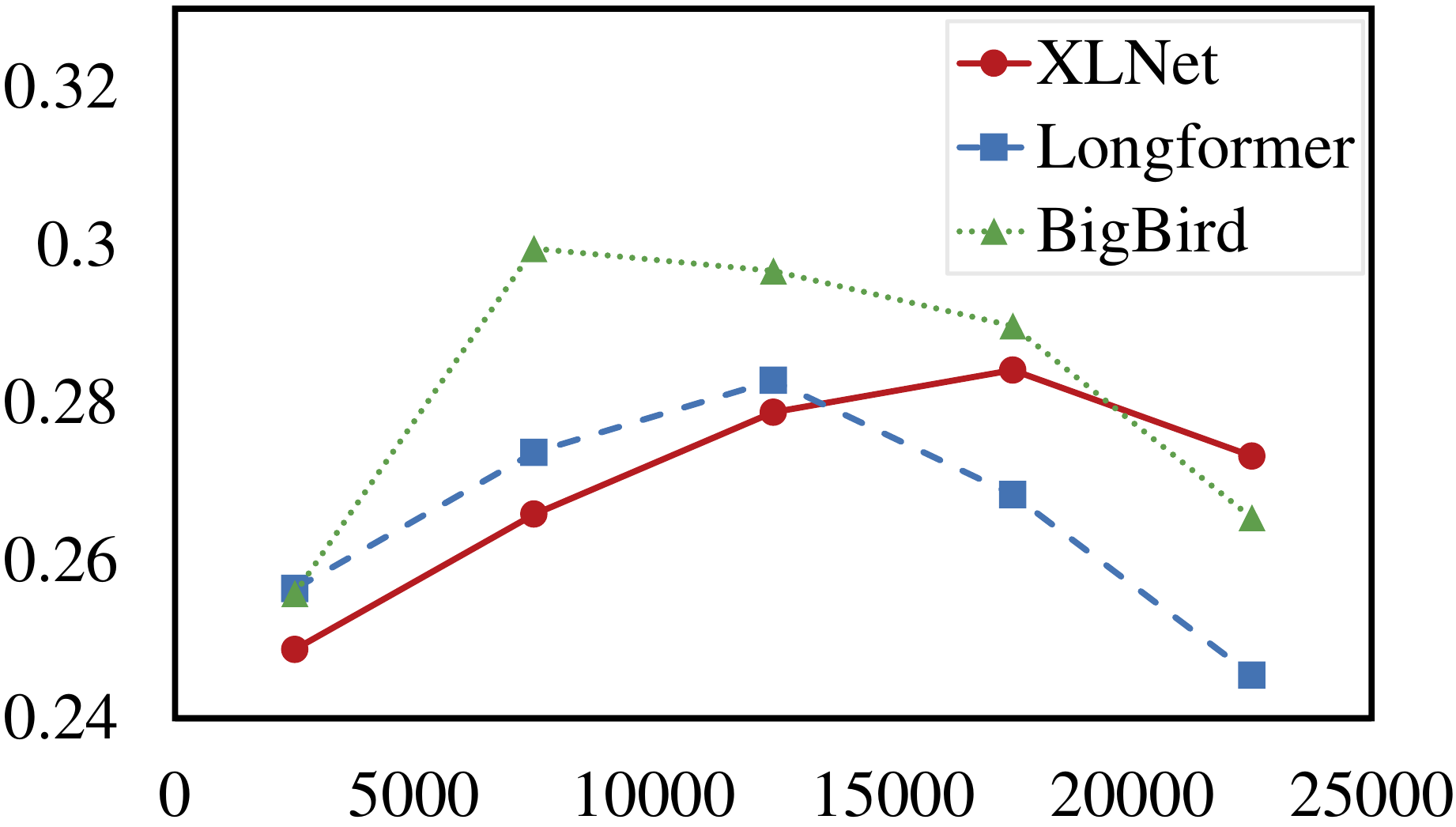}}
	\caption{Performance ($y$ axis) on DLUE datasets with different document lengths ($x$ axis).}
	\label{doc_len}
\end{figure*}

\subsection{Overall Results}

Table~\ref{overall_result} shows the overall results on DLUE. From this table, we can see that:

\textbf{1) Document understanding is far from being solved.} 
From Table~\ref{overall_result}, we can see that the best benchmark system can only achieve 44.5 average score. While it's difficult to establish an accurate human performance ceiling in DLUE, we can take some indicators to prove that the performance gap between human and models are huge. For example, human agreement on ECOM was measured at 80.8\% F1~\citep{xu2022eco}, much higher than our best baseline of 39.1\% F1. Likewise, ~\citet{dasigi-etal-2021-dataset} study a subset of Qasper that has multiple annotated answers, and find their overlap to be 60.9\% F1, more than double our best baseline. This indicates that contemporary off-the-shelf models struggle with documents, challenging future work to make progress on DLUE.

\textbf{2) Different tasks have different advantageous architectures, raising a need for an universal document understanding architecture which can dominate all tasks in one architecture.}
From Table~\ref{overall_result}, we can see that different model architectures seem to be good at processing different tasks.
Specifically, the performance of XLNet ranks first on the structure analysis tasks, while Longformer and BigBird perform better on the other tasks. Linformer and Performer do well on document classification tasks. This shows that recurrence-based models may have advantages over hierarchically structured data and pattern-based models may be more effective on flat data. Contrary to the other tasks, fast low-rank/kernel-based models do better on document classification tasks.
No architecture dominates all tasks, which indicates that more universal models are needed.

\textbf{3) Lengthy content is the critical, but not the only, challenge for document understanding.}
From Table~\ref{overall_result} and Table~\ref{dataset}, we can see that models perform poorly with too long inputs, such as the 18.5 best $F_1$ score in NarrativeQA dataset with 51790 average input length. However, even for those structure analysis and extraction tasks where documents can be taken in completely by long-range transformer models, the model performances still fail to meet expectations. Obviously, there exist other challenges for document understanding apart from lengthy input, such as complex structures and dispersed knowledge.

\textbf{4) It is critical to take global context into consideration.} 
From Table~\ref{overall_result}, we can see that long-range transformers that can take in more contexts achieve a higher score than vanilla transformer in most datasets.
This demonstrates longer contexts are necessary to understand documents. Document-level tasks can't be solved in the same way as short-text tasks.

\subsection{Computational Efficiency}

The last column of Table~\ref{overall_result} shows inference speeds of models. For a fair comparison, we use the standard test datasets of DLUE as testbed. Based on our implementation, the low-rank/kernel-based models are the fastest. Performer model is the fastest model with 6.7 steps per second, which is close to the inference speed of Linformer with 6.4 steps per second. The results are consistent with model complexity, which has a significant impact on inference speed. The low-rank/ kernel-based models decompose the $N\times N$ self-attention matrix to a lower-dimensional representation and thus usually have a $O(N)$ time complexity. Pattern-based models sparsify the attention matrix according to predefined or learnable patterns and the time complexity is usually between $O(N)$ and $O(N^2)$. Recurrence-based models connect multiple segments and blocks via recurrence and the representative XLNet has a $O(N^2)$ time complexity.

\subsection{Effect Of Document Length}

To investigate how the document length will impact the performance, we cluster the documents into buckets for each task according to their document lengths and run the evaluation on each bucket. The breakdown analysis is shown in Figure \ref{doc_len}.

On the whole, understanding longer documents faces more challenges.
We notice that the performances on most datasets decrease when document lengths increase, with ContractNLI dataset as an exception. This maybe because there exists label bias related to document lengths in ContractNLI datasets. We find that a longer contract tends to entail a hypothesis, with 34\% probability for documents shorter than 1000 words and 76\% probability for documents longer than 5000 words.

The performance of pattern-based models seems to be more stable when the document lengths increase. We can see that Longformer and BigBird obtain a greater advantage when documents get longer. We think there are two reasons. First, the global token mechanism in Longformer and BigBird could help models focus on important information and be less distracted by noise in long contexts. Second, the maximum input length of XLNet is smaller due to the segment-level recurrence mechanism.

The performance is relatively stable on datasets where document lengths far exceed input limits. Figure \ref{fig:subfig:f} shows performance on NarrativeQA dataset. When the document length exceeds 20,000 tokens, the result remains around 18 $F_1$ for Longformer, BigBird and 15 $F_1$ for XLNet. This indicates the ability of efficient transformers to understand long documents are limited. 

\subsection{Effect of dispersed knowledge Exploition}

Our goal in this section is to validate that recognizing and aggregating dispersed knowledge is crucial to document understanding, and there is still much room for current models to improve. We analyze from two perspectives: 1) the effect of mention distance, which can be viewed as the measure of dispersion; 2) performance comparison between long-range transformers and short-text models without global information.

\begin{figure} [t!]
	\centering
	\setlength{\belowcaptionskip}{-0.4cm}
	\includegraphics[width=0.4\textwidth]{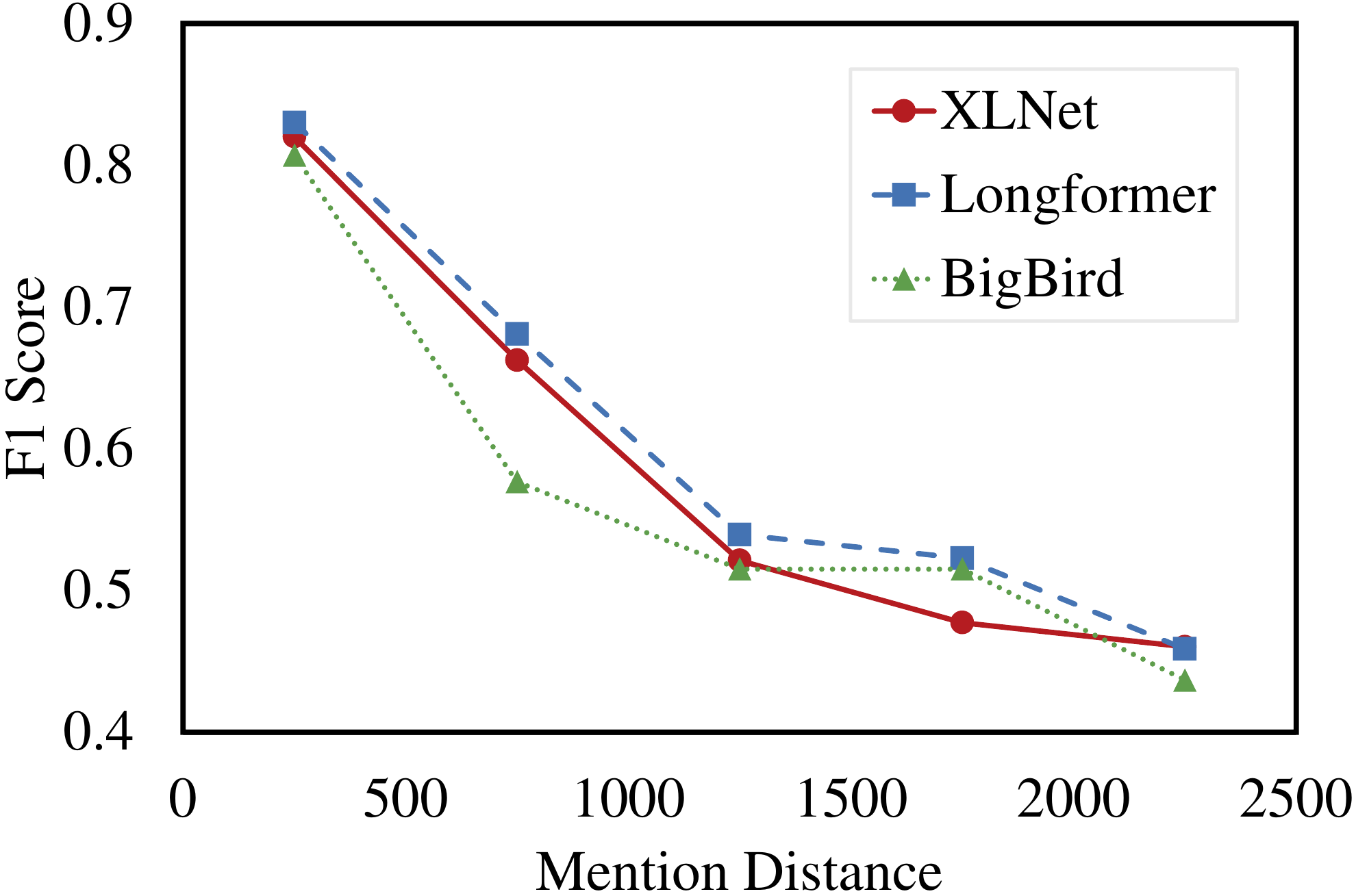}
	\caption{Performance on LitBank dataset with different mention distances. Mention distances can reflect the degree of knowledge dispersion in a document.}
	\label{mention_distance}
\end{figure}

\textbf{Effect of Mention Distance}
To quantify the impact of dispersed knowledge to document understanding, we analyze the performance of coreference resolution with different mention distances on LitBank dataset.
From Figure \ref{mention_distance}, we can see that the performance of all models decreases sharply when the mention distances increase. This indicates long-distance coreference is more challenging than within-sentence coreference. It's easy to understand because it puts forward higher requirements for the ability to capture and aggregate dispersed information.
We can also notice the huge performance gap between short and long mention distances, which indicates there is still much room for further improvements in models' ability of integrating global information.

\textbf{Comparison with Short-text Models}
To verify the importance of global information to document understanding, we compare the performance of long-range transformers with two existing short-text models, including CogLTX~\citep{ding2020cogltx} and ToBERT~\citep{pappagari2019hierarchical}. CogLTX jointly trains two BERT models to select key sentences from documents. ToBERT divides documents into smaller chunks and uses a transformer layer over BERT-based chunk representations. We select Hyperpartisan and Qasper datasets, whose tasks can be solved by CogLTX and ToBERT, and in which documents can be completely taken in by long-range transformers to eliminate interference caused by more contexts. 

From Table~\ref{short_text_model_result}, we can see that long-range transformers do have advantages over IR-based and chunking-based methods. 
Intuitively, the reason behind is that the performance of long-range transformers benefits from the contextual representation with a broader view of the document.
These findings emphasize the need for future studies in document understanding to integrate global information.
The results also indicate that DLUE effectively covers the assessment of the ability to recognize and aggregate dispersed knowledge across the whole text.

\begin{table}[!t]
    \centering
    \setlength{\belowcaptionskip}{-0.5cm}
    \renewcommand{\arraystretch}{0.8}
    \resizebox{0.4\textwidth}{!}{
        \begin{tabular}{ccc}
            \toprule
            Model & Hyperpartisan & Qasper \\
            \midrule
            XLNet &  81.4 & 24.2\\
            Longformer & 83.8 & \textbf{26.4}\\
            BigBird & \textbf{85.9} & 26.2\\
            \midrule
            CogLTX & 82.9 & 18.9\\
            ToBERT & 78.4 & 16.6\\
            \bottomrule
        \end{tabular}
    }
    \caption{Performance comparison between long-range transformers and short-text models.}
    \label{short_text_model_result}
\end{table}

\section{Conclusions}

We propose a new benchmark DLUE that places the spot on documents and their lengthy content, complex underlying structure and dispersed knowledge challenges. DLUE covers diverse document-level tasks to evaluate four basic abilities required by document understanding, including document classification, document structure analysis, document information extraction and document transcription. Based on DLUE, we conduct an extensive side-by-side comparison of three document understanding architectures. Experiments demonstrate document understanding is far from being solved, and there exists a need for a universal architecture that can dominate all tasks.

\section*{Limitations}
DLUE now focuses on plain text documents, while the documents one encounter, e.g., scientific articles, company announcements, or even personal notes, may also contain multi-modal information and with non-sequential structure. In future work, we intend to integrate these multi-modal, complex structure information into our document understanding benchmark.

Besides, due to the huge cost of computing resources, we didn't pretrain transformer models specialized for document understanding, but directly use the public pretrained versions or train from scratch. We believe an unified pretraining by also incorporating document-related tasks will further enhance the understanding performance.

\section*{Ethics Statement}
In consideration of ethical concerns, we provide the following detailed description:
\begin{enumerate}
    \item We believe that this work is beneficial to develop universal document understanding architectures, which can help people quickly get information from business documents, legal statements and so on, saving time and money.
    \item We standardize and put together ten datasets, which are all already publicly available under CC-BY-(NC-)SA-4.0 licenses\footnote{https://creativecommons.org/licenses/by/4.0/}. For all the datasets, we have referenced the original work and encouraged DLUE users to do so.
    \item All DLUE benchmark datasets have low ethical risks and do not expose any sensitive or personally identifiable information.
\end{enumerate}

\bibliography{anthology,custom}
\bibliographystyle{acl_natbib}

\end{document}